\documentclass[10pt,twocolumn,letterpaper]{article}

\usepackage{iccv}              

\definecolor{iccvblue}{rgb}{0.21,0.49,0.74}
\usepackage[pagebackref,breaklinks,colorlinks,allcolors=iccvblue]{hyperref}

\usepackage[utf8]{inputenc}
\usepackage{amsmath, amssymb, amsthm}
\usepackage{algorithm}
\usepackage{algpseudocode}
\usepackage{comment}
\usepackage{tikz}
\usetikzlibrary{shadows,arrows.meta, positioning, fit, shapes.geometric, calc}
\usepackage{pgfplots}
\pgfplotsset{compat=1.17} 
\usepgfplotslibrary{groupplots}
\usepackage{fontawesome5} 
\usepackage{graphicx}
\usepackage{hyperref}
\usepackage{enumitem}
\usepackage{caption}
\usepackage{lipsum}
\usepackage{multirow}

\tikzstyle{block} = [rectangle, draw, rounded corners, text width=7em, text centered, minimum height=3em]
\tikzstyle{decision} = [diamond, draw, aspect=2, text width=4em, text centered, inner sep=0pt]
\tikzstyle{arrow} = [thick,->,>=stealth]

\def\paperID{15418} 
\def\confName{ICCV}
\def\confYear{2025}

\title{FAIR-SIGHT: Fairness Assurance in Image Recognition via Simultaneous Conformal Thresholding and Dynamic Output Repair}

\author{Arya Fayyazi\\
University of Southern California\\
Los Angeles, California, USA\\
{\tt\small afayyazi@usc.edu}
\and
Mehdi Kamal\\
University of Southern California\\
Los Angeles, California, USA\\
{\tt\small mehdi.kamal@usc.edu}
\and
Massoud Pedram\\
University of Southern California\\
Los Angeles, California, USA\\
{\tt\small pedram@usc.edu}
}

\begin{document}
\maketitle
\begin{abstract}
We introduce FAIR-SIGHT, an innovative post-hoc framework designed to ensure fairness in computer vision systems by combining conformal prediction with a dynamic output repair mechanism. Our approach calculates a fairness-aware non-conformity score that simultaneously assesses prediction errors and fairness violations. Using conformal prediction, we establish an adaptive threshold that provides rigorous finite-sample, distribution-free guarantees. When the non-conformity score for a new image exceeds the calibrated threshold, FAIR-SIGHT implements targeted corrective adjustments, such as logit shifts for classification and confidence recalibration for detection, to reduce both group and individual fairness disparities—-all without the need for retraining or having access to internal model parameters. Comprehensive theoretical analysis validates our method's error control and convergence properties. At the same time, extensive empirical evaluations on benchmark datasets show that FAIR-SIGHT significantly reduces fairness disparities while preserving high predictive performance.
\end{abstract}
    
\section{Introduction}
\label{sec:introduction}

Advances in deep learning~\cite{10.1145/3658617.3697777,10.1145/3658617.3697547,azizi2025mambaextend} have propelled computer vision systems to near-human performance in tasks such as object detection, semantic segmentation, and face recognition~\cite{He2016Deep,zhang2020celeba,movahhedrad2024green}. As these systems are increasingly integrated into high-stakes applications, from security and autonomous driving to healthcare, the risk of unequal treatment across demographic groups has garnered significant attention~\cite{Sabouri2025}. The early revelations of biases in facial analysis~\cite{buolamwini2018gender} underscored the potential for serious social harm, prompting a proliferation of research on fairness in vision~\cite{xu2023fair,lee2023survey}. However, most existing solutions modify model architectures or retrain models with fairness constraints~\cite{madras2018learning,roh2020fairbatch,kamiran2012data}, approaches that are often impractical for proprietary large-scale systems or models that are computationally expensive to retrain. Moreover, static fairness regularizers embedded at training time may fail to adapt as data distributions shift, gradually eroding fairness guarantees.

Motivated by these limitations, we present a new perspective on the enforcement of fairness in computer vision (CV), one that requires \emph{no retraining} or internal parameter access but still offers \emph{formal statistical guarantees} to limit unfair outcomes. Our approach draws on recent extensions of \emph{conformal prediction} (CP)~\cite{shafer2008tutorial,angelopoulos2023conformal,fayyazi2025facter}, a distribution-free framework that provides guarantees of finite sample coverage under exchangeability assumptions. In previous work, FACTER~\cite{fayyazi2025facter} demonstrated the viability of CP for fairness in recommendation systems; here, we extend those ideas to high-dimensional vision outputs such as bounding boxes and pixel-level predictions.

\begin{figure}[t]
\centering
\begin{tikzpicture}[font=\sffamily, 
    node distance=1.8cm, 
    >=stealth,
    line width=0.5pt,
    scale=0.45, 
    transform shape
]

\tikzstyle{stage} = [
    rectangle,
    rounded corners,
    align=center,
    drop shadow,
    fill=blue!5,
    draw=blue!70,
    text width=2.8cm,
    minimum height=2.0cm
]
\tikzstyle{stage2} = [
    rectangle,
    rounded corners,
    align=center,
    drop shadow,
    fill=red!5,
    draw=red!70,
    text width=2.8cm,
    minimum height=2.0cm
]
\tikzstyle{stage3} = [
    rectangle,
    rounded corners,
    align=center,
    drop shadow,
    fill=green!5,
    draw=green!70,
    text width=2.8cm,
    minimum height=2.0cm
]
\tikzstyle{stage4} = [
    rectangle,
    rounded corners,
    align=center,
    drop shadow,
    fill=orange!5,
    draw=orange!70,
    text width=2.8cm,
    minimum height=2.0cm
]
\tikzstyle{arrow} = [->, thick, color=gray!70]

\node[stage](input){
    \Large\faCamera\\[3pt]
    \textbf{Images}\\
    \small(Protected Attribute)
};
\node[stage2, right=of input](model){
    \Large\faRobot\\[3pt]
    \textbf{Black-Box}\\
    \small{CV Model}
};
\node[stage3, right=of model](faircp){
    \Large\faIcon{shield-alt}\\[3pt]
    \textbf{FAIR-SIGHT}\\
    \small{Post-hoc Repair}
};
\node[stage4, right=of faircp](output){
    \Large\faSmile\\[3pt]
    \textbf{Fair Output}\\
    \small{Less Bias \&\\Better Results}
};

\draw[arrow] (input) -- (model) 
    node[midway, above, font=\footnotesize, color=gray!80]{\faArrowRight};
\draw[arrow] (model) -- (faircp) 
    node[midway, above, font=\footnotesize, color=gray!80]{\faArrowRight};
\draw[arrow] (faircp) -- (output) 
    node[midway, above, font=\footnotesize, color=gray!80]{\faArrowRight};

\node[font=\Large, color=blue!50] at ($(input.north) + (0,1.0)$) { \faImages };
\node[font=\Large, color=red!50] at ($(model.north) + (0,1.0)$) { \faBrain };
\node[font=\Large, color=green!50] at ($(faircp.north) + (0,1.0)$) { \faBalanceScale };
\node[font=\Large, color=orange!50] at ($(output.north) + (0,1.0)$) { \faHeart };

\end{tikzpicture}

\caption{High-level overview of proposed workflow. We start with input images (possibly containing a protected attribute), feed them into a black-box computer vision (CV) model, then apply \textbf{FAIR-SIGHT} module as a post-hoc fairness repair. Its output is a \emph{fair} set of predictions, mitigating bias while preserving accuracy.}
\label{fig:teaser_diagram}
\end{figure}
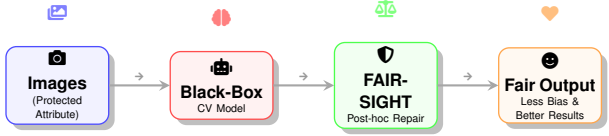

Our framework, \textbf{FAIR-SIGHT} (Figure~\ref{fig:teaser_diagram}), tackles the dual challenge of (i) maintaining high precision of the vision model and (ii) mitigating biases at both the group level and the individual level. We achieve this by combining CP's interpretability and statistical coverage properties with an adjustable, task-driven penalty function that quantifies fairness deviations. Crucially, this \emph{post hoc} strategy (1) takes a black-box model as given, (2) calibrates fairness thresholds on a held-out dataset, and (3) applies real-time repairs (e.g., logit shifts for classification, confidence recalibration for detection) when predictions exceed the calibrated threshold. By adopting a dynamic, data-driven thresholding mechanism, our approach adapts to evolving distributions, thereby addressing the shortcomings of static fairness interventions.

Beyond its practical advantages, FAIR-SIGHT establishes formal fairness guarantees by ensuring that the fraction of outputs flagged as unfair can be controlled by a user-specified significance level \(\alpha\); that is, at most \(\alpha\) of future samples are expected to exceed our fairness threshold. This statistical coverage is particularly valuable in complex scenarios—such as object detection or instance segmentation—where standard metrics like IoU or AP must be carefully balanced across protected groups.

In this paper, we show how to formalize fairness deficits as non-conformity scores that integrate both predictive error and demographic disparity, calibrate these scores via conformal prediction to yield robust fairness thresholds, and apply dynamic repairs with an adaptive feedback loop such that fairness constraints hold even as data distributions shift. Our experiments indicate that FAIR-SIGHT consistently reduces group-level disparities and enhances individual-level consistency in classification and detection tasks while preserving high accuracy.


\section{Preliminaries}
\label{sec:preliminaries}

In this section, we set the stage for our proposed \textbf{FAIR-SIGHT} framework by reviewing advances in fairness-aware computer vision, formalizing core fairness concepts, and laying out the central ideas of conformal prediction. This background contextualizes the methodological details in Section~\ref{sec:method}.

\subsection{Related Works}
\label{subsec:related}

Fairness in computer vision has garnered increased attention as vision-based systems are increasingly deployed in sensitive areas such as healthcare, surveillance, and autonomous driving. Early studies revealed demographic biases in facial analysis \cite{buolamwini2018gender}, driving further investigation into biases across a range of tasks, including object detection, semantic segmentation, and image captioning. Below, we summarize recent contributions that reflect the fast-evolving state of fairness-aware vision research.
Recent studies have sought to ensure equitable bounding-box predictions across demographic groups. For instance, Xu et al.~\cite{zhang2021fairmot} introduced fairness-aware detectors that dynamically align confidence thresholds for protected and non-protected groups, reducing performance gaps in person detection. These methods often require modifying architectures or retraining from scratch, posing challenges for black-box or proprietary models.

Much work has also addressed pixel-level tasks. Lee et al.~\cite{lee2023survey} integrated fairness constraints into semantic segmentation losses, underscoring the tension between enforcing parity in pixel-level predictions and sustaining accuracy. Such approaches frequently embed fairness regularizers during training, limiting post-deployment adaptivity.

As data distributions shift over time, static fairness interventions risk losing efficacy. A dynamic threshold adaptation scheme was presented in~\cite{wang2023fairness} that periodically recalibrates decision criteria to maintain demographic parity. Although effective, these methods typically require partial retraining or consistent updates to internal model parameters.

Post-hoc fairness modules have emerged as a practical solution to accommodate scenarios where model internals are inaccessible. Dubey et al.~\cite{dubey2024nested} explored external calibration pipelines that wrap around black-box classifiers to adjust skewed outputs. However, these methods often lack finite-sample coverage guarantees, and their fairness improvements can be degraded when data distributions evolve.

Conformal prediction (CP) has recently been extended to ensure statistical coverage in complex tasks such as object segmentation and multi-label recognition~\cite{stutts2023lightweight}. Although CP has long been used to provide classification uncertainty sets, the use of CP for fairness is a more recent development. FACTER~\cite{fayyazi2025facter} demonstrated how CP-based thresholds can control fairness violation rates in the language-model-driven recommendation.
\vspace{-0.5em}

\paragraph{Our Contribution.}
\vspace{-0.5em}

Despite the advances cited above, a practical gap remains: post-hoc fairness methods that require neither architectural modifications nor retraining often lack rigorous distribution-free guarantees, whereas conformal-based solutions have seldom targeted comprehensive fairness in high-dimensional vision tasks. Building on the insights of dynamic calibration~\cite{wang2023fairness}, external repair modules~\cite{dubey2024nested}, and CP-based coverage~\cite{stutts2023lightweight,fayyazi2025facter}, our framework unifies these concepts to deliver finite sample fairness guarantees in both classification and detection, without internal model access. In doing so, we address the typical limitations of prior work, namely, reliance on retraining and the absence of formal coverage bounds, offering a robust post hoc strategy for fairness in diverse CV tasks.

\subsection{Fairness Definitions}
\label{subsec:fairness-defs}

\paragraph{Group Fairness.}
Group fairness requires that performance metrics be comparable across subpopulations \cite{binns2020apparent}. If $\mathcal{G}_0$ and $\mathcal{G}_1$ denote non-protected and protected groups, then for a metric $\text{Metric}(\cdot)$,
\[
\bigl|\text{Metric}(\mathcal{G}_0) - \text{Metric}(\mathcal{G}_1)\bigr| \;\le\; \epsilon,
\]
for some small $\epsilon>0$. 

\paragraph{Individual Fairness.}
Individual fairness posits that \emph{similar inputs yield similar outputs} \cite{petersen2021post}. Under \emph{minimal attribute change fairness}, flipping the protected attribute $A$ from 0 to 1 should not drastically alter the prediction:
\[
\bigl\| f(I, A=0) - f(I, A=1)\bigr\| \;\le\;\delta,
\]
for a small $\delta>0$.
\subsection{Problem Statement}
\label{subsec:prob-statement}

Let $I\in\mathbb{R}^{H\times W\times3}$ be an image and $f: I\to Y$ a trained (but black-box) model, e.g.~a classifier or detector. Each $I$ has a protected attribute label $A\in\{0,1\}$. Our \emph{post-hoc}, model-agnostic objective is to adjust $f(I)$ for new images so that:
\begin{enumerate}[label=(\roman*), leftmargin=1.2em]
    \item \textbf{Group Fairness} holds, keeping group-level metric disparities below $\epsilon$.
    \item \textbf{Individual Fairness} holds, preventing large output changes from minimal $A$-flips.
\end{enumerate}
We aim to accomplish this \emph{without} altering the internal weights of $f$. Key challenges include measuring fairness violations in complex outputs, calibrating thresholds that separate fair vs.\ unfair predictions, and adaptively repairing them online.

\subsection{Introduction to Conformal Prediction}
\label{subsec:conformal-bg}

Conformal prediction~\cite{shafer2008tutorial} provides guarantees of finite-sample coverage under exchangeability. By mapping each sample $(I_i, A_i)$ to a \emph{non-conformity score} $S(I_i)$ and sorting these scores on a calibration set, one obtains a threshold $Q_{\alpha}$ for a chosen significance level $\alpha$. The probability that a new sample score $S(I_{\mathrm{new}})$ exceeds $Q_{\alpha}$ is at most $\alpha$, i.e.:
\[
\Pr\bigl[S(I_{\mathrm{new}}) > Q_{\alpha}\bigr]\;\le\;\alpha.
\]
This property naturally supports \emph{fairness control}: if $S(I)$ encodes fairness violations, upper bounding $S(I)$ by $Q_{\alpha}$ ensures that no more than the $\alpha$-percentage of future samples will exhibit unfairness.


The details of implementing and algorithmically realizing these ideas follow in Section~\ref{sec:method}, where we formally describe \emph{FAIR-SIGHT} and analyze its theoretical guarantees.

\section{Methodology and Algorithm}
\label{sec:method}

\begin{figure}[h]
\centering
\begin{tikzpicture}[
    font=\footnotesize,
    node distance=0.6cm,
    auto,
    >=latex,
    scale=0.75,
    transform shape,
    block/.style={
        rectangle,
        draw=blue!60,
        fill=blue!10,
        thick,
        rounded corners,
        text width=5em,
        align=center,
        minimum height=2.5em
    },
    blockwide/.style={
        rectangle,
        draw=blue!60,
        fill=blue!10,
        thick,
        rounded corners,
        text width=5.5em,
        align=center,
        minimum height=2.5em
    },
    onblock/.style={
        rectangle,
        draw=green!60,
        fill=green!10,
        thick,
        rounded corners,
        text width=5em,
        align=center,
        minimum height=2.5em
    },
    decision/.style={
        diamond,
        aspect=2,
        draw=orange!80,
        fill=orange!10,
        thick,
        align=center,
        text width=2em,
        inner sep=1pt
    },
    database/.style={
        cylinder,
        shape border rotate=90,
        draw,
        fill=yellow!20,
        aspect=0.2,
        text width=2.5em,
        minimum height=3em,
        align=center
    },
    cloud/.style={
        ellipse,
        draw=red!80,
        fill=red!10,
        thick,
        align=center,
        minimum width=3em,
        minimum height=2em
    },
    line/.style={draw, thick, ->}
]

\node[blockwide, fill=blue!30] (offlineTitle) {Offline Calibration};
\node[blockwide, below=0.5cm of offlineTitle] (dataset) {Calib. Data \\ \((I_i,A_i)\)};
\node[blockwide, below=0.5cm of dataset] (preprocess) {Preprocess \& \\ Features};
\node[blockwide, below=0.5cm of preprocess] (scoreCalc) {Non-conf. \\ Scores \\ \((S(I_i)\,\text{or}\,S_R)\)};
\node[blockwide, below=0.5cm of scoreCalc] (quantile) {Calibrate \\ \(Q_\alpha\)};
\node[database, below=0.5cm of quantile] (threshStore) {Store};

\draw [line] (offlineTitle) -- (dataset);
\draw [line] (dataset) -- (preprocess);
\draw [line] (preprocess) -- (scoreCalc);
\draw [line] (scoreCalc) -- (quantile);
\draw [line] (quantile) -- (threshStore);

\node[block, below left=0.2cm and 0.1cm of scoreCalc, text width=5em, fill=blue!5] (detectAnnot)
  {Detect: partition \(\{R_j\}\)};
\node[block, below right=0.2cm and 0.1cm of scoreCalc, text width=5em, fill=blue!5] (classAnnot)
  {Classif.: logit/emb.};

\draw [line, dashed] (detectAnnot.north) |- (scoreCalc.west);
\draw [line, dashed] (classAnnot.north) |- (scoreCalc.east);

\node[onblock, fill=green!30, right=3.1cm of offlineTitle] (onlineTitle) {Online Inference};
\node[onblock, below=0.5cm of onlineTitle] (newImage) {New Image \\ \(I_{\text{new}}\)};
\node[onblock, below=0.5cm of newImage] (model) {CV Model};
\node[onblock, below=0.5cm of model, text width=5.5em] (scoreNew) {Compute \\ \(S(I_{\text{new}})\)};
\node[decision, below=0.5cm of scoreNew] (decide) {\(S > Q?\)};
\node[cloud, below left=0.6cm and 0.2cm of decide, text width=5em] (correct) {Correct \\ (logit/bbox)};
\node[onblock, below=0.6cm of correct, text width=5em] (update) {Update \\ \(Q^{(t+1)}\)};
\node[onblock, below right=0.6cm and 0.2cm of decide, text width=5em] (output) {Output};

\draw [line] (onlineTitle) -- (newImage);
\draw [line] (newImage) -- (model);
\draw [line] (model) -- (scoreNew);
\draw [line] (scoreNew) -- (decide);
\draw [line] (decide) -| node[left, sloped]{Yes} (correct);
\draw [line] (decide) -| node[left, sloped]{No} (output);
\draw [line] (correct) -- (update);
\draw [line] (update) -- (output);

\draw [line, dashed] (update.west) -- ++(-1.0,0) |- node[midway, fill=white]{Feedback} (threshStore.east);

\node[block, fill=green!5, text width=4.5em, above right=0.2cm and 0.1cm of model] (bbox) {Bbox + Conf.};
\draw [->, dashed] (bbox.south) -- (model.north);
\node[block, fill=green!5, text width=4.5em, above left=0.2cm and 0.1cm of model] (prot) {Prot. Attr. \\ \(A=1\)};
\draw [->, dashed] (prot.south) -- (model.north);

\end{tikzpicture}
\caption{\textbf{FAIR-SIGHT Workflow.} The \textbf{Offline Calibration} (left) takes a calibration dataset \((I_i, A_i)\), processes each sample to compute non-conformity scores (\(S(I_i)\) or region-based \(S_R(I_i,R_j)\)), and derives the conformal fairness threshold(s) \(\{Q_\alpha,\,Q_\alpha(R_j)\}\). These thresholds are stored for later use. The \textbf{Online Inference} (right) processes each new image \(I_{\text{new}}\) through the trained computer vision (CV) model, computes \(S(I_{\text{new}})\), and checks it against the stored thresholds. If \(S(I_{\text{new}})\) (or any \(S_R(I_{\text{new}},R_j)\)) exceeds the threshold, we apply a \emph{post hoc} correction (e.g., adjusting class logits or bounding-box confidences) and optionally \emph{update} the threshold through the feedback loop. Otherwise, the raw model output is used as-is.}
\label{fig:cv_FAIR-SIGHT_workflow_compact}
\vspace{-1em}
\end{figure}
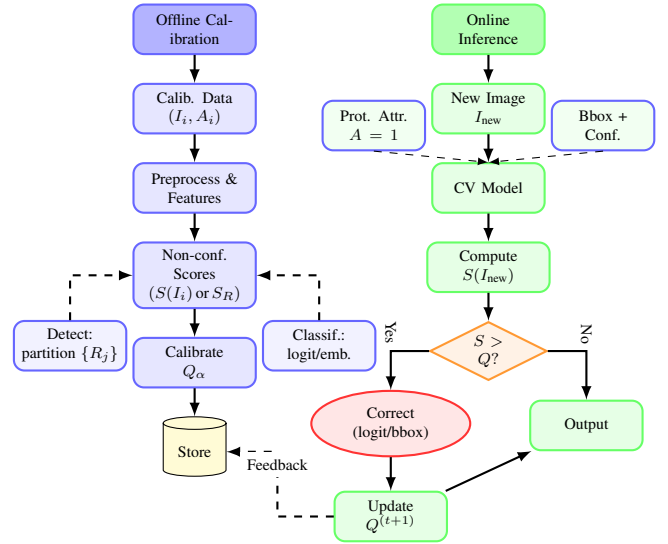

\textbf{FAIR-SIGHT} (as shown in Figure~\ref{fig:cv_FAIR-SIGHT_workflow_compact}) is a framework that combines conformal prediction with dynamic post-hoc fairness repair methods. Hence, FAIR-SIGHT enforces rigorous fairness criteria in classification and detection tasks without retraining or modifying the internal parameters of a model.

\subsection{Overall Problem Setup}
\label{subsec:method-setup}

\paragraph{Inputs and Protected Attributes.}
We consider an image $I\in\mathbb{R}^{H\times W\times3}$ accompanied by a binary protected attribute $A\in\{0,1\}$. Typically, $A=1$ denotes membership in an underrepresented demographic group. We aim to ensure that the model predictions do not systematically disadvantage images with $A=1$.
\vspace{-1em}

\paragraph{Outputs.}
FAIR-SIGHT assesses potential biases in the outputs of classification and detection tasks and applies repairs when necessary. The output in these two tasks is defined by the following:
\begin{itemize}[leftmargin=1em]
    \item \textbf{Classification.} A black-box model $f$ produces a logit/probability vector $\ell \in \mathbb{R}^K$, with the final label given by $\hat{y}=\arg\max_k \ell_k$. The model does not use $A$ during inference.
    \item \textbf{Detection.} A black-box model $g$ yields bounding boxes $\{\mathbf{b}_i\}$ with the corresponding confidence scores $\{s_i\}$ (and possibly class labels $\{\ell_i\}$). Again, $A$ is not used by $g$.
\end{itemize}

\subsection{Fairness-Aware Non-Conformity Scores}
\label{subsec:nonconf-score}

Central to our approach is a non-conformity score $S(I)$ that merges predictive error with a fairness penalty:
\begin{equation}
\label{eq:nonconf-expanded}
S(I) \;=\; d\bigl(h(I),\,y_{\text{ref}}(I)\bigr)\;+\;\lambda\,\Delta(I,A),
\end{equation}

\noindent where, $h(I)$ represents the raw output of the model (for example, logits for classification, detection confidences for detection) and $y_{\text{ref}}(I)$ denotes the reference or the ground truth output. $d(\cdot,\cdot)$ measures predictive error (e.g.~$1-\mathrm{softmax}_{\text{true}}$ for classification or $1-\mathrm{mIoU}$ for detection). $\Delta(I,A)$ is a fairness penalty that quantifies how much $h(I)$ deviates from group-fair behavior. For example, in detection, if images with $A=1$ consistently have lower bounding-box confidences, $\Delta(I,A)$ becomes large. And finally, $\lambda>0$ balances the fairness penalty relative to predictive error.

For detection tasks, we further partition each image into spatial regions $\{R_j\}$ (e.g., uniform grid) and define a regional non-conformity score $S_R(I, R_j)$, which can capture localized fairness violations. The final score $S(I)$ may be an aggregate (e.g., sum or max) of the region-level scores.


\subsection{Offline Calibration via Conformal Prediction}
\label{subsec:method-calibration}

We assemble a calibration set $\mathcal{D}_{\mathrm{cal}}=\{(I_i,A_i,y_i)\}_{i=1}^n$ that is distinct from the training data. For each image $I_i$, we compute the non-conformity score $S(I_i)$ (and, if applicable, each regional score $S_R(I_i,R_j)$). Sorting these scores, we define the conformal threshold:
\begin{equation}
\label{eq:qalpha}
Q_\alpha \;=\;\inf\!\Bigl\{
q : \frac{1}{n+1}\sum_{i=1}^n \mathbb{I}\{S(I_i)\le q\} \ge 1-\alpha
\Bigr\}.
\end{equation}
Under the assumption of exchangeability, this threshold guarantees that for a new image $I_{\mathrm{new}}$,
\[
\Pr\{ S(I_{\mathrm{new}}) > Q_\alpha \} \le \alpha.
\]
For detection tasks, region-specific thresholds $Q_\alpha(R_j)$ are computed similarly from the scores $\{S_R(I_i,R_j)\}$.

\subsection{Online Inference and Fairness Repair}
\label{subsec:method-online}

During inference, each new image $(I_{\mathrm{new}}, A_{\mathrm{new}})$ is processed by the vision model to obtain a raw output $h(I_{\mathrm{new}})$, and its non-conformity score $S(I_{\mathrm{new}})$ is computed. If $S(I_{\mathrm{new}}) \le Q_\alpha$ (and, for detection, all regional scores satisfy $S_R(I_{\mathrm{new}},R_j) \le Q_\alpha(R_j)$), the result is accepted as fair. Otherwise, a repair mechanism is activated.

\vspace{0.5em}
\noindent\textbf{Classification Repair.} We adjust the logit corresponding to the true class by adding a constant correction term for classification tasks. Concretely, if a sample violates the fairness threshold ($S(I_{\mathrm{new}}) > Q_\alpha$), we compute
\[
\Delta_c \;=\;\min\Bigl\{\kappa\,\bigl(S(I_{\mathrm{new}})-Q_\alpha\bigr),\,\Delta_{\max}\Bigr\},
\]
where $\kappa>0$ is a scaling factor and $\Delta_{\max}$ is an upper bound preventing overcorrection. These constants are chosen based on cross-validation on the calibration set (see our ablation subsection for details), ensuring that the corrected logit mitigates the fairness penalty without unduly distorting the model’s confidence.

\vspace{0.5em}
\noindent\textbf{Detection Repair.} In detection tasks, bounding-box confidence scores for the protected group ($A=1$) may be scaled by a factor $\eta$ if they fall below the calibrated threshold. We select $\eta$ from a discrete set of candidate values (e.g., below 1.0 for reducing overconfident boxes or above 1.0 for boosting underconfident ones) according to performance on the calibration set. This process is neither random nor uniform; rather, we systematically evaluate fairness metrics (e.g., AP and AP Gap) under different $\eta$ and pick the best setting that best balances between accuracy and reduced disparities. Full details appear in our ablation study.

\vspace{0.5em}
\noindent\textbf{Adaptive Threshold Update.} If repeated fairness violations persist after repair, an optional update rule refines the threshold:
\[
Q_\alpha^{(t+1)} \;=\;\gamma\,Q_\alpha^{(t)} 
\;+\;
(1-\gamma)\,\min\bigl\{Q_\alpha^{(t)},\,S(I_{\mathrm{new}})\bigr\}.
\]
Here, $\gamma\in(0,1)$ is a decay factor determined via hyperparameter tuning. As shown in our ablation subsection, this mechanism incrementally tightens the threshold when the system encounters multiple high non-conformity scores, enforcing stricter fairness constraints over time.

\vspace{0.3em}
Algorithm~\ref{alg:fvcp} summarizes the overall procedure. Importantly, all repairs and threshold updates occur post hoc at the output level, leaving the underlying model parameters unchanged.

\begin{algorithm}[t]
\caption{FAIR-SIGHT: Offline Calibration + Online Fairness Enforcement}
\label{alg:fvcp}
\begin{algorithmic}[1]
\Require 
$\mathcal{D}_{\mathrm{cal}}$, significance $\alpha$, fairness weight $\lambda$, (optional) threshold update rate $\gamma$, region bins $\{R_j\}$ (for detection)
\Statex
\State \textbf{Offline: Conformal Calibration}
\For{$i=1,\ldots,n$}
    \State $S(I_i) \leftarrow d\bigl(h(I_i),y_{\mathrm{ref}}(I_i)\bigr) + \lambda\,\Delta(I_i,A_i)$
    \If{detection task}
        \For{each region $R_j$ in $I_i$}
            \State $S_R(I_i,R_j) \leftarrow \dots$  \quad // defined over spatial regions (e.g., uniform grid)
        \EndFor
    \EndIf
\EndFor
\State $Q_\alpha \leftarrow \text{Quantile}(\{S(I_i)\}, 1-\alpha)$
\If{detection task}
    \For{each region $R_j$}
        \State $Q_\alpha(R_j) \leftarrow \text{Quantile}(\{S_R(I_i,R_j)\}, 1-\alpha)$
    \EndFor
\EndIf
\Statex
\State \textbf{Online: Inference and Repair}
\For{\textbf{each} new image $(I_{\mathrm{new}},A_{\mathrm{new}})$}
    \State $S_{\mathrm{new}} \leftarrow d\bigl(h(I_{\mathrm{new}}),y_{\mathrm{ref}}(I_{\mathrm{new}})\bigr) + \lambda\,\Delta(I_{\mathrm{new}},A_{\mathrm{new}})$
    \If{$S_{\mathrm{new}} \le Q_\alpha$ \textbf{and} (for detection: $S_R(I_{\mathrm{new}},R_j)\le Q_\alpha(R_j)$ for all $R_j$)}
        \State \textbf{Output} $h(I_{\mathrm{new}})$
    \Else
        \State $\hat{y} \leftarrow \mathrm{Repair}\Bigl(h(I_{\mathrm{new}}), S_{\mathrm{new}}, Q_\alpha\Bigr)$ \quad // e.g., logit shift or score scaling
        \If{adaptive threshold update is enabled}
            \State $Q_\alpha \leftarrow \gamma\,Q_\alpha + (1-\gamma)\,\min\{Q_\alpha, S_{\mathrm{new}}\}$
        \EndIf
        \State \textbf{Output} $\hat{y}$
    \EndIf
\EndFor
\end{algorithmic}
\end{algorithm}

\subsection{Key Theoretical Insights and Guarantees}
\label{subsec:method-theory}

This subsection outlines the formal properties that underlie our approach, demonstrating how \textbf{FAIR-SIGHT} leverages conformal prediction to control fairness violations under realistic conditions, maintains robustness against small output perturbations, and dynamically adapts thresholds without retraining.

\vspace{-0.5em}
\paragraph{Finite-Sample Fairness Coverage.}
Conformal prediction \cite{shafer2008tutorial} provides a powerful guarantee: when non-conformity scores \(\{S(I_i)\}_{i=1}^n\) are computed on a calibration set and sorted in non-decreasing order, selecting the \(\lceil (n+1)(1-\alpha)\rceil\)-th score as \(Q_\alpha\) ensures, for any future image \(I_{\mathrm{new}}\),
\begin{equation*}
\Pr\bigl[S(I_{\mathrm{new}}) \;\le\;Q_\alpha\bigr] 
\;\;\ge\;1-\alpha.
\end{equation*}
Because each non-conformity score \(S(I)\) encodes both accuracy-related error and a fairness penalty, surpassing \(Q_\alpha\) indicates a \emph{fairness violation}. Hence, at most, an \(\alpha\)-fraction of new samples have scores above \(Q_\alpha\), bounding the fraction of unfair outcomes. This sets an explicit, data-driven limit on fairness violations in a \emph{finite-sample} and \emph{distribution-free} manner—particularly relevant for high-dimensional vision tasks where conventional analytical bounds may fail.

\vspace{-0.5em}
\paragraph{Robustness under Lipschitz Continuity.}
Let the model output for image \(I\) be \(h(I)\), and let each component of \(S(I)\) (the predictive error \(d(\cdot)\) and fairness penalty \(\Delta(\cdot)\)) be Lipschitz in \(\|h(I_1)-h(I_2)\|\). Concretely, if there exist constants \(L_d\) and \(L_\Delta\) such that
\[
\bigl|\,d\bigl(h(I_1)\bigr) - d\bigl(h(I_2)\bigr)\bigr|
\;\le\;L_d\,\bigl\|\,h(I_1)-h(I_2)\bigr\|,
\]
\[
\bigl|\Delta(I_1,A_1) - \Delta(I_2,A_2)\bigr|
\;\le\;L_\Delta\,\bigl\|\,h(I_1)-h(I_2)\bigr\|,
\]
then the overall score \(S(I)\) is Lipschitz with constant \((L_d + \lambda\,L_\Delta)\). Consequently, small fluctuations in the logit or detection confidences of the model produce only minor changes in \(S(I)\). This ensures that borderline samples near \(Q_\alpha\) remain stable against minor noise and fosters robustness when fairness thresholds are applied in the real world. 

\vspace{-0.5em}
\paragraph{Adaptive Thresholding and No-Retraining Requirement.}
Unlike adversarial or reweighting methods that must retrain the entire model, \textbf{FAIR-SIGHT} modifies \emph{only} the raw output if the non-conformity score exceeds \(Q_\alpha\). This design is both \emph{post hoc} and \emph{model-agnostic}, which do not require parameter-level access. In addition, as explained previously, we can optionally employ an update rule for \(Q_\alpha\) based on repeated violations.
Recurrent violations lead \(Q_\alpha\) to decrease, imposing more stringent fairness constraints. Under mild boundedness assumptions, this update converges to a fixed point that reflects the observed distribution of scores, further fortifying fairness coverage as data distributions shift over time. By avoiding retraining, \textbf{FAIR-SIGHT} remains viable for industrial black-box models and can swiftly recalibrate its threshold to maintain formal fairness guarantees under changing conditions.

\subsection{Key Contributions and Achievements}
\label{subsec:achievements}

Our FAIR-SIGHT framework brings several key advancements, including 1) it unifies fairness control for both classification and detection without retraining the underlying model, 2) by embedding fairness penalties into a non-conformity score and calibrating an adaptive threshold via conformal prediction, our method provides formal, finite-sample guarantees on the rate of fairness violations, and 3) in detection tasks, our region-based thresholding enables localized repairs, ensuring that spatial disparities are addressed precisely. 

\vspace{0.5em}
\textbf{Note on Formulation:}  
In our framework, the fairness penalty term $\Delta(I,A)$ captures the discrepancy between the model's output for an image with a given protected attribute and similar images with an alternative attribute. In detection, spatial regions $R_j$ are defined using a uniform grid over the image, though alternative segmentation methods can be employed. The repair mechanisms, such as adding a constant to logits or scaling confidence scores, are chosen based on empirical studies and theoretical intuition that small, targeted corrections can effectively reduce fairness violations.


\section{Results and Discussion}
\label{sec:experiments}
\vspace{-0.5em}

In this section, we present a comprehensive evaluation of our \textbf{FAIR-SIGHT} framework on both classification and detection tasks. We begin by describing the baseline methods, backbone architectures, and datasets used (\S\ref{subsec:compared-methods}), then detail our evaluation metrics, hyperparameter settings (including an ablation study), and system configuration (\S\ref{subsec:eval-and-impl}). Finally, we report quantitative (\S\ref{subsec:results-classification}, \S\ref{subsec:results-detection}) and qualitative (\S\ref{subsec:qualitative-results}) results, accompanied by a broader discussion of limitations and potential failure modes (\S\ref{subsec:discussion-limitations}).
\vspace{-0.5em}

\subsection{Considered Methods, Models, and Datasets}
\label{subsec:compared-methods}
\vspace{-0.5em}

\paragraph{Methods:}
We compare the efficacy of the \textbf{FAIR-SIGHT} with those of three methods, including:
\begin{itemize}[leftmargin=1em]
    \item \textbf{Baseline:} The unmodified vision model, which does not incorporate any fairness constraints.
    \item \textbf{AdvDebias:} An adversarial fairness approach \cite{madras2018learning}, training an additional adversarial branch to minimize representation of the protected attribute in the learned features. This often requires expensive retraining.
    \item \textbf{FairBatch:} A reweighting strategy \cite{roh2020fairbatch,kamiran2012data} that adjusts the sampling distribution during training to reduce demographic disparities. Although effective, it also involves retraining and is less adaptable to postdeployment changes.
\end{itemize}
Compared to the AdvDebias and FairBatch methods, \textbf{FAIR-SIGHT} is post hoc, requiring no retraining or parameter-level access, and remains adaptable via its conformal calibration and dynamic repair.

\textbf{Model Backbones:}
We evaluate four representative architectures, including ResNet50 \cite{He2016Deep}, ResNet101 \cite{He2016Deep}, MambaVision-T-1K \cite{hatamizadeh2024mambavision}, and MambaVision-L2-1K \cite{hatamizadeh2024mambavision}.
Transformer-based MambaVision backbones generally yield higher accuracy than ResNets. The L2 variant outperforms T-1K, reflecting architectural depth and training scale differences.

\textbf{Datasets:}
We consider three datasets in the evaluation studies: CelebA, UTKFace, and COCO.
\textbf{CelebA} \cite{zhang2020celeba} is a large-scale face attribute dataset. We predict the \texttt{Smiling} attribute while treating \texttt{Female} as the protected attribute ($A=1$ for female).
\textbf{UTKFace} \cite{zhang2017age} contains more than 20k facial images labeled with age, gender, and ethnicity. We formulate a binary age classification task (\texttt{Young} vs.\ \texttt{Not Young}) and designate the \textit{Black} ethnicity as protected ($A=1$). 

For object detection, we use a \textbf{COCO}-based subset \cite{lin2014microsoft} focusing on the \texttt{person} class. Because COCO does not provide demographic labels, we apply FairFace checkpoints \cite{karkkainen2021fairface} to infer each individual's race. We again designate \textit{Black} ($A=1$) as the protected group. Inferring attributes in this manner can introduce label noise (e.g., partially obscured faces may be incorrectly identified). Despite these limitations, we find this approach sufficient for showing our method’s robustness; in real-world deployments, more rigorous validation or human auditing of protected-attribute labels would be recommended.

\subsection{Evaluation Metrics and Implementation Details}
\label{subsec:eval-and-impl}

\paragraph{Classification Metrics:}
We measure \textbf{Accuracy} and \textbf{AUC} (Area Under the ROC Curve) to assess predictive performance. Also, we extract the \textbf{DPD} (Demographic Parity Difference) and \textbf{EOD} (Equalized Odds Difference) \cite{hardt2016equality} metrics, where the smaller value of these metrics indicates better fairness. Finally, group-specific \textbf{TPR} metric is obtained to reveal if one group receives systematically different rates of correct predictions.

\textbf{Detection Metrics:}
Following standard COCO evaluation \cite{lin2014microsoft}, we extract the \textbf{AP(prot)} and \textbf{AP(nonprot)} metrics, which show the average precision for the protected and non-protected groups, respectively. Also, to measure the group fairness improvement in the detection task, we use \textbf{Gap} metric (Gap = AP(nonprot) $-$ AP(prot)).

\textbf{System Configuration.}
We implemented all experiments in Python 3.8 using PyTorch 2.1 on an 8-GPU NVIDIA RTX A6000 server (CUDA 12.4). For AdvDebias and FairBatch, we retrain from ImageNet-pretrained weights using Adam (learning rate $1\times10^{-4}$) for 10--20 epochs. By contrast, FAIR-SIGHT only uses a 80\% calibration set (20\% for testing) to compute non-conformity scores and thresholds, then applies repairs at inference—avoiding any need to re-engineer or retrain the underlying model. In practical deployments, reliance on labeled protected attributes for calibration can be a limitation if such labels are scarce or if fairness definitions evolve (e.g., intersectional or multi-attribute fairness). However, we find it sufficient for these benchmark tasks.

\subsection{Hyperparameters and Ablation Study}
\label{subsec:ablation}

FAIR-SIGHT relies on a small set of hyperparameters, divided into (i) \emph{conformal calibration} and (ii) \emph{repair mechanisms}. The calibration parameters include the significance level \(\alpha\) and the decay factor \(\gamma\), controlling how strictly the threshold \(Q_\alpha\) is enforced and tightened over time. The repair parameters include \(\lambda\) (balancing predictive error vs. fairness penalties), \(\eta\) (scaling under- or overconfident outputs), \(\kappa\) (dictating how strongly logits are shifted for classification) and \(\Delta_{\max}\) (preventing excessive correction).

\begin{figure*}[t]
\centering
\begin{tikzpicture}
\begin{groupplot}[
    group style={
        group size=4 by 1,
        horizontal sep=1.2cm,
    },
    width=3.8cm,
    height=4.0cm,
    xlabel style={font=\small},
    ylabel style={font=\small},
    tick label style={font=\scriptsize},
    legend style={font=\scriptsize, at={(6,1.0)}, anchor=north},
]

\nextgroupplot[
    xlabel={$\lambda$},
    ylabel={Metric Value},
    xtick={0.1,0.3,0.5,0.7,0.9,1.1},
    ymin=0, ymax=0.25,
]
\addplot[color=blue, mark=square, thick] coordinates {(0.1,0.180) (0.3,0.155) (0.5,0.140) (0.7,0.115) (0.9,0.125) (1.1,0.135)};
\addlegendentry{DPD};
\addplot[color=red, mark=triangle, thick] coordinates {(0.1,0.100) (0.3,0.085) (0.5,0.070) (0.7,0.055) (0.9,0.065) (1.1,0.075)};
\addlegendentry{EOD};

\nextgroupplot[
    xlabel={$\gamma$},
    xtick={0.85,0.90,0.95,0.99},
    ymin=0, ymax=0.15,
]
\addplot[color=blue, mark=square, thick] coordinates {(0.85,0.130) (0.90,0.120) (0.95,0.115) (0.99,0.118)};
\addplot[color=red, mark=triangle, thick] coordinates {(0.85,0.085) (0.90,0.075) (0.95,0.070) (0.99,0.074)};

\nextgroupplot[
    xlabel={$\eta$},
    xtick={0.6,0.8,1.0,1.2},
    ymin=0, ymax=0.20,
]
\addplot[color=blue, mark=square, thick] coordinates {(0.6,0.140) (0.8,0.115) (1.0,0.120) (1.2,0.125)};
\addplot[color=red, mark=triangle, thick] coordinates {(0.6,0.075) (0.8,0.055) (1.0,0.060) (1.2,0.065)};

\nextgroupplot[
    xlabel={$\kappa$},
    xtick={0.2,0.5,0.8,1.0},
    ymin=0, ymax=0.25,
]
\addplot[color=blue, mark=square, thick] coordinates {(0.2,0.190) (0.5,0.160) (0.8,0.120) (1.0,0.125)};
\addplot[color=red, mark=triangle, thick] coordinates {(0.2,0.105) (0.5,0.080) (0.8,0.060) (1.0,0.065)};

\end{groupplot}
\end{tikzpicture}
\vspace{-1em}
\caption{Ablation study on FAIR-SIGHT hyperparameters (ResNet50, CelebA). Each panel plots fairness metrics (DPD/EOD) against a different parameter: \(\lambda\) (error vs.\ fairness penalty), \(\gamma\) (threshold update), \(\eta\) (scaling factor), and \(\kappa\) (logit shift aggressiveness). Middle-range values minimize disparities without harming accuracy.}
\label{fig:ablation}
\end{figure*}
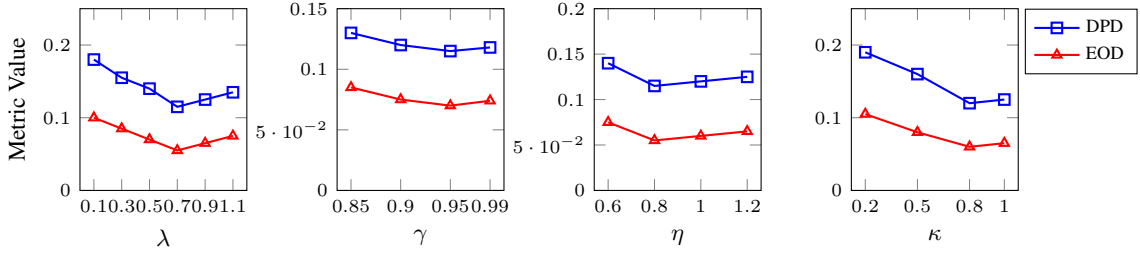

\noindent
Figure~\ref{fig:ablation} illustrates how each parameter influences Demographic Parity Difference (DPD) and Equalized Odds Difference (EOD). In all cases, moderate settings minimize group-level disparities while preserving accuracy. We adopt these empirically validated configurations throughout our experiments (Section~\ref{sec:experiments}), ensuring a stable trade-off between fairness and predictive performance.

\subsection{Classification Results}
\label{subsec:results-classification}

Tables~\ref{tab:celeba-results} and \ref{tab:utkface-results} summarize classification outcomes on CelebA and UTKFace, respectively. While accuracy and AUC are comparable across methods, \textbf{FAIR-SIGHT} substantially reduces fairness disparities: up to 30\% lower DPD and EOD relative to the baseline. Notably, AdvDebias and FairBatch also improve fairness but demand retraining and may be less adaptable when data shift post-deployment.

\begin{table}[t]
\centering
\caption{\textbf{CelebA Classification Results.} FAIR-SIGHT reduces DPD/EOD by over 30\% compared to Baseline while maintaining high accuracy and AUC.}
\label{tab:celeba-results}
\resizebox{\linewidth}{!}{%
\begin{tabular}{llccccc}
\toprule
\textbf{Backbone} & \textbf{Method} & \textbf{Accuracy} & \textbf{DPD} & \textbf{EOD} & \textbf{AUC} & \textbf{G0 TPR / G1 TPR}\\
\midrule
\multirow{4}{*}{\emph{ResNet50 \cite{He2016Deep}}}
  & Baseline       & 0.915 & 0.160 & 0.080 & 0.917 & 0.865 / 0.925 \\
  & AdvDebias      & 0.908 & 0.130 & 0.070 & 0.910 & 0.870 / 0.920 \\
  & FairBatch      & 0.911 & 0.125 & 0.067 & 0.913 & 0.870 / 0.930 \\
  & \textbf{FAIR-SIGHT} & \textbf{0.918} & \textbf{0.115} & \textbf{0.065} & \textbf{0.919} & \textbf{0.875 / 0.940}\\
\midrule
\multirow{4}{*}{\emph{ResNet101 \cite{He2016Deep}}}
  & Baseline       & 0.924 & 0.155 & 0.075 & 0.923 & 0.885 / 0.940 \\
  & AdvDebias      & 0.915 & 0.120 & 0.067 & 0.917 & 0.880 / 0.920 \\
  & FairBatch      & 0.920 & 0.110 & 0.065 & 0.918 & 0.890 / 0.930 \\
  & \textbf{FAIR-SIGHT} & \textbf{0.926} & \textbf{0.095} & \textbf{0.055} & \textbf{0.926} & \textbf{0.895 / 0.945}\\
\midrule
\multirow{4}{*}{\emph{MambaVision-T-1K \cite{hatamizadeh2024mambavision}}}
  & Baseline       & 0.935 & 0.140 & 0.060 & 0.934 & 0.895 / 0.945 \\
  & AdvDebias      & 0.928 & 0.110 & 0.050 & 0.930 & 0.885 / 0.940 \\
  & FairBatch      & 0.931 & 0.105 & 0.048 & 0.932 & 0.890 / 0.945 \\
  & \textbf{FAIR-SIGHT} & \textbf{0.940} & \textbf{0.085} & \textbf{0.040} & \textbf{0.939} & \textbf{0.900 / 0.950}\\
\midrule
\multirow{4}{*}{\emph{MambaVision-L2-1K \cite{hatamizadeh2024mambavision}}}
  & Baseline       & 0.940 & 0.135 & 0.055 & 0.939 & 0.900 / 0.950 \\
  & AdvDebias      & 0.934 & 0.105 & 0.045 & 0.936 & 0.895 / 0.940 \\
  & FairBatch      & 0.936 & 0.100 & 0.043 & 0.937 & 0.900 / 0.945 \\
  & \textbf{FAIR-SIGHT} & \textbf{0.943} & \textbf{0.080} & \textbf{0.035} & \textbf{0.942} & \textbf{0.905 / 0.955}\\
\bottomrule
\end{tabular}}
\end{table}

\vspace{-0.5em}
\begin{table}[t]
\centering
\vspace{-0.5em}
\caption{\textbf{UTKFace Classification Results.} FAIR-SIGHT lowers DPD/EOD by over 25\% relative to Baseline, while slightly improving accuracy and AUC.}
\label{tab:utkface-results}
\resizebox{\linewidth}{!}{%
\begin{tabular}{llccccc}
\toprule
\textbf{Backbone} & \textbf{Method} & \textbf{Accuracy} & \textbf{DPD} & \textbf{EOD} & \textbf{AUC} & \textbf{G0 TPR / G1 TPR} \\
\midrule
\multirow{4}{*}{\emph{ResNet50 \cite{He2016Deep}}}
  & Baseline      & 0.800 & 0.210 & 0.100 & 0.802 & 0.780 / 0.875 \\
  & AdvDebias     & 0.790 & 0.160 & 0.085 & 0.795 & 0.770 / 0.860 \\
  & FairBatch     & 0.804 & 0.155 & 0.080 & 0.805 & 0.790 / 0.865 \\
  & \textbf{FAIR-SIGHT} & \textbf{0.815} & \textbf{0.135} & \textbf{0.065} & \textbf{0.810} & \textbf{0.800 / 0.885} \\
\midrule
\multirow{4}{*}{\emph{ResNet101 \cite{He2016Deep}}}
  & Baseline      & 0.815 & 0.205 & 0.095 & 0.817 & 0.790 / 0.880 \\
  & AdvDebias     & 0.800 & 0.160 & 0.085 & 0.804 & 0.770 / 0.860 \\
  & FairBatch     & 0.810 & 0.155 & 0.080 & 0.815 & 0.785 / 0.870 \\
  & \textbf{FAIR-SIGHT} & \textbf{0.825} & \textbf{0.140} & \textbf{0.070} & \textbf{0.822} & \textbf{0.800 / 0.890} \\
\midrule
\multirow{4}{*}{\emph{MambaVision-T-1K \cite{hatamizadeh2024mambavision}}}
  & Baseline      & 0.845 & 0.185 & 0.090 & 0.840 & 0.815 / 0.900 \\
  & AdvDebias     & 0.835 & 0.155 & 0.080 & 0.833 & 0.805 / 0.880 \\
  & FairBatch     & 0.850 & 0.150 & 0.078 & 0.848 & 0.815 / 0.895 \\
  & \textbf{FAIR-SIGHT} & \textbf{0.860} & \textbf{0.135} & \textbf{0.065} & \textbf{0.857} & \textbf{0.825 / 0.910} \\
\midrule
\multirow{4}{*}{\emph{MambaVision-L2-1K \cite{hatamizadeh2024mambavision}}}
  & Baseline      & 0.860 & 0.175 & 0.085 & 0.858 & 0.830 / 0.915 \\
  & AdvDebias     & 0.850 & 0.145 & 0.075 & 0.847 & 0.820 / 0.900 \\
  & FairBatch     & 0.865 & 0.140 & 0.072 & 0.863 & 0.830 / 0.905 \\
  & \textbf{FAIR-SIGHT} & \textbf{0.875} & \textbf{0.125} & \textbf{0.065} & \textbf{0.872} & \textbf{0.840 / 0.920} \\
\bottomrule
\end{tabular}%
}
\end{table}

\subsection{Detection Results}
\label{subsec:results-detection}

Table~\ref{tab:improved_detection_table} shows detection performance on a COCO-based subset. Because MambaVision models are more advanced, their AP values are higher overall. \textbf{FAIR-SIGHT} consistently boosts AP for $A=1$ (the protected group) and narrows the AP Gap compared to baselines, indicating more equitable performance. Moreover, FAIR-SIGHT is computationally efficient as it operates in a post hoc manner without requiring retraining, unlike AdvDebias and FairBatch, which require additional training and inference overhead. Although the protected attribute labels derived from FairFace~\cite{karkkainen2021fairface} can be noisy (e.g. due to partial occlusion), this does not prevent our method from achieving significant fairness improvements while maintaining high detection quality.

\begin{table}[t]
\centering
\caption{\textbf{Detection Results on COCO-based Subset.} AP(prot) and AP(nonprot) measure performance for protected vs.\ non-protected groups; Gap = AP(nonprot) $-$ AP(prot). Lower Gap shows improved fairness, higher AP indicates stronger detection.}
\label{tab:improved_detection_table}
\resizebox{\linewidth}{!}{%
\begin{tabular}{lcccc}
\toprule
\textbf{Backbone} & \textbf{Method} & \textbf{AP(prot)} & \textbf{AP(nonprot)} & \textbf{Gap} \\
\midrule
\multirow{4}{*}{ResNet50 \cite{He2016Deep}} 
& Baseline    & 0.532 & 0.603 & 0.071 \\
& AdvDebias   & 0.510 & 0.575 & 0.065 \\
& FairBatch   & 0.556 & 0.600 & 0.044 \\
& \textbf{FAIR-SIGHT} & \textbf{0.577} & \textbf{0.610} & \textbf{0.033} \\
\midrule
\multirow{4}{*}{ResNet101 \cite{He2016Deep}} 
& Baseline    & 0.541 & 0.586 & 0.045 \\
& AdvDebias   & 0.523 & 0.562 & 0.039 \\
& FairBatch   & 0.547 & 0.586 & 0.040 \\
& \textbf{FAIR-SIGHT} & \textbf{0.551} & \textbf{0.568} & \textbf{0.017} \\
\midrule
\multirow{4}{*}{MambaVision-T-1K \cite{hatamizadeh2024mambavision}} 
& Baseline    & 0.620 & 0.663 & 0.043 \\
& AdvDebias   & 0.590 & 0.621 & 0.031 \\
& FairBatch   & 0.625 & 0.651 & 0.026 \\
& \textbf{FAIR-SIGHT} & \textbf{0.660} & \textbf{0.669} & \textbf{0.009} \\
\midrule
\multirow{4}{*}{MambaVision-L2-1K \cite{hatamizadeh2024mambavision}} 
& Baseline    & 0.623 & 0.665 & 0.042 \\
& AdvDebias   & 0.610 & 0.640 & 0.030 \\
& FairBatch   & 0.656 & 0.691 & 0.035 \\
& \textbf{FAIR-SIGHT} & \textbf{0.702} & \textbf{0.710} & \textbf{0.008} \\
\bottomrule
\end{tabular}}
\end{table}

\subsection{Qualitative and Visualization Results}
\label{subsec:qualitative-results}

\begin{figure}[htbp]
\centering
\begin{tikzpicture}[scale=0.6]
  \node (tsneLeft) at (0,0) {
    \includegraphics[width=4.0cm]{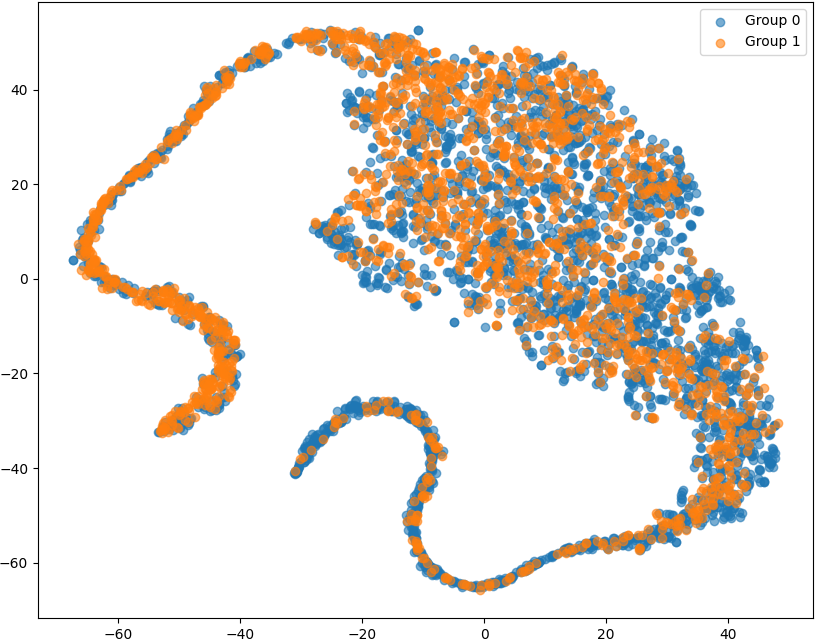}
  };
  \node (tsneRight) at (7,0) {
    \includegraphics[width=4.0cm]{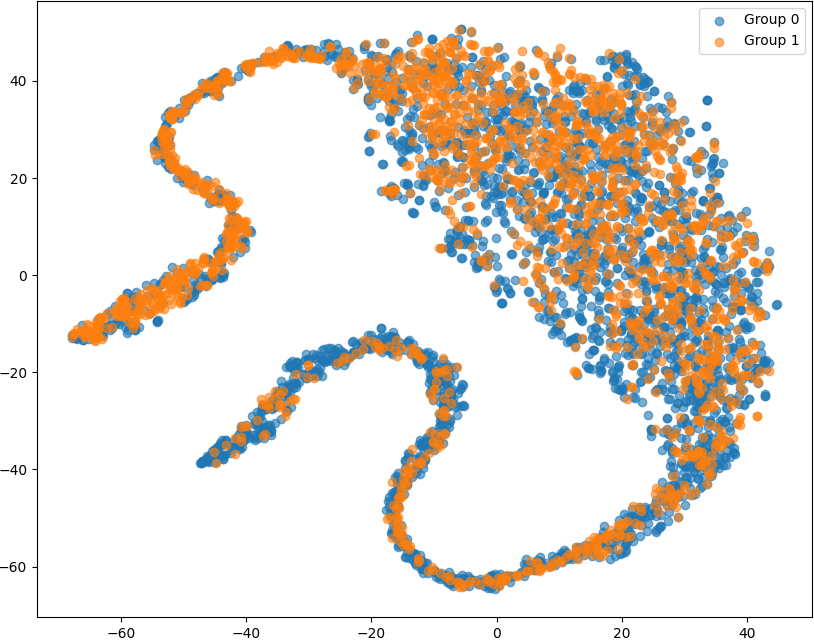}
  };
  \coordinate (midTop) at ($(tsneLeft.north)!0.5!(tsneRight.north)$);
  \coordinate (arrowTopLeft) at ($(midTop)+(-1.0,0)$);
  \coordinate (arrowTopRight) at ($(midTop)+(1.0,0)$);
  \draw[->, thick] (arrowTopLeft) -- (arrowTopRight)
    node[midway, above]{\footnotesize Intermingled Clusters};
\end{tikzpicture}

\vspace{0.3cm}

\begin{tikzpicture}[scale=0.6]
  \node (detLeft) at (0,0) {
    \includegraphics[width=4cm]{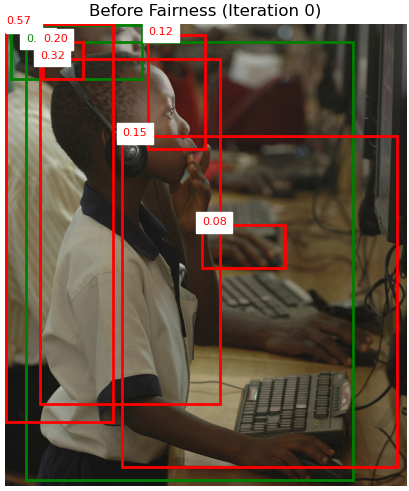}
  };
  \node (detRight) at (7,0) {
    \includegraphics[width=4cm]{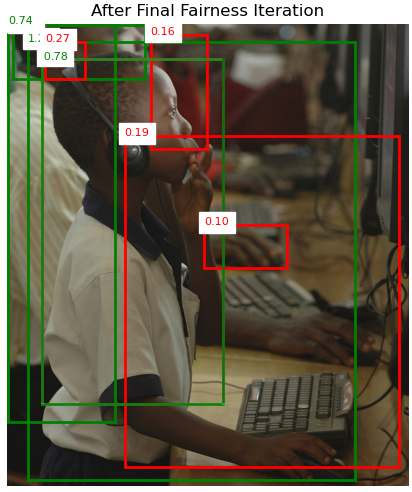}
  };
  \coordinate (midTop2) at ($(detLeft.north)!0.5!(detRight.north)$);
  \coordinate (arrowTopLeft2) at ($(midTop2)+(-1.0,0)$);
  \coordinate (arrowTopRight2) at ($(midTop2)+(1.0,0)$);
  \draw[->, thick] (arrowTopLeft2) -- (arrowTopRight2)
    node[midway, above]{\footnotesize Confidence Repair};
\end{tikzpicture}

\caption{\textbf{Qualitative Results.} \emph{Top:} t-SNE embeddings on UTKFace classification illustrate that baseline features cluster by protected attribute, whereas FAIR-SIGHT produces more intermingled clusters, indicating reduced bias in the feature space. \emph{Bottom:} On the COCO-based detection dataset using a MambaVision-L2-1K model, a conformal calibration threshold of 0.73 is computed from the validation set. In the baseline output (left), several bounding boxes for persons in the protected group (race = Black) have confidence scores below 0.73 (indicated by red boxes), signaling under-detection for the protected group. After applying FAIR-SIGHT’s post-hoc repair mechanism, the scores of those protected group boxes are boosted so that detections meet the threshold, resulting in more balanced and fair outputs (right).}
\label{fig:combined}
\end{figure}
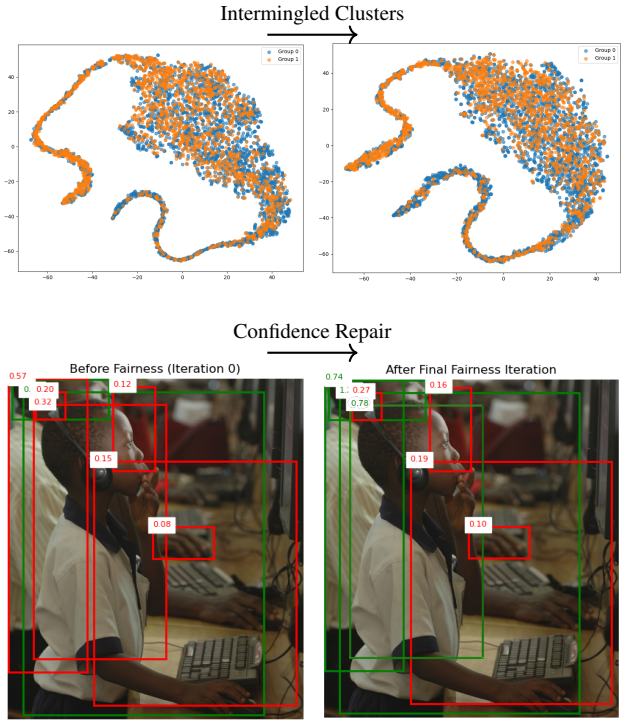

Figure~\ref{fig:combined} (top) shows t-SNE \cite{van2008visualizing} embeddings on UTKFace using the MambaVision-L2-1K, where group 1 (Black persons) and group 0 (other races) are distinguished. The baseline exhibits noticeable clustering by protected attribute, suggesting feature separation that can lead to biased predictions. Under FAIR-SIGHT, these clusters become more intermixed, implying a reduction in group-specific feature alignment and translating into lower DPD/EOD. Figure~\ref{fig:combined} (bottom) illustrates how bounding-box confidences for the protected group are boosted or scaled as necessary, reducing the AP Gap between the two groups in detection tasks.

\subsection{Merits and Limitations of FAIR-SIGHT }
\label{subsec:discussion-limitations}
\vspace{-0.3em}

\paragraph{+ Post-hoc Strategy.}
While adversarial debiasing (AdvDebias) and sample reweighting (FairBatch) can also enhance fairness, these methods typically demand retraining, which is computationally costly, and are less adaptable if data distributions shift. FAIR-SIGHT, by contrast, applies a \emph{post hoc} mechanism using conformal prediction, providing direct control over the fraction of permissible fairness violations (\(\alpha\)) and obviating expensive retraining steps.
\vspace{-0.5em}

\paragraph{+ Generalizability of Repairs.}
Although our repair mechanisms are illustrated primarily for classification (via logit shifts) and detection (via bounding-box confidence scaling), the underlying approach can be generalized to other computer vision tasks, such as semantic segmentation or multi-modal outputs (e.g., image-caption pairs). In those settings, one would define appropriate non-conformity scores and repair functions that adjust relevant output dimensions (e.g., pixel-level predictions for segmentation) consistent with group/individual fairness. The core conformal thresholding procedure remains unchanged, suggesting that FAIR-SIGHT could be extended to a broad range of vision applications where outputs are structured, high-dimensional, or multi-modal.
\vspace{-0.5em}

\paragraph{- Label Noise and Protected Attribute Availability.}
Our detection experiments infer protected attributes from FairFace checkpoints~\cite{karkkainen2021fairface}, which may introduce label noise if faces are partially visible or occluded. Although we observe robust improvements despite this potential noise, real-world deployments should, where possible, ensure more reliable demographic labeling (e.g., manual audits or advanced face attribute estimators). Moreover, FAIR-SIGHT relies on a calibration set with known protected attributes, which can be challenging if the user environment lacks explicit demographic data or if fairness definitions (e.g., intersectional or multi-attribute) evolve over time.

\vspace{-1em}

\paragraph{- Exchangeability and Multi-Attribute Fairness.}
FAIR-SIGHT’s conformal guarantees rely on an assumption of exchangeability. In practice, non-stationary or correlated data (e.g., seasonal domain shifts and intersectional attributes) may challenge this assumption, potentially weakening coverage guarantees. While we focus on a single binary attribute, real-world fairness often involves multiple or intersecting attributes (e.g., gender \emph{and} age \emph{and} ethnicity). Extending FAIR-SIGHT to such settings would require more intricate penalty definitions and calibration schemes, an important direction for future research.
\vspace{-1em}
\paragraph{- Feature Space vs.\ Prediction Bias.}
Our t-SNE analysis shows that intermingled embeddings correlate with reduced group disparities in predictions. When features from different groups reside in shared clusters, the model is less prone to group-specific biases in classification or bounding-box assignment. Nevertheless, a thoroughly intermixed feature space does not guarantee perfect fairness, nor is partial separation always unfair. Our findings highlight the empirical connection between feature alignment and fairness, though deeper theoretical investigation could refine this relationship.

\vspace{-0.5em}

\section{Conclusion}
\label{sec:conclusion}

We presented FAIR-SIGHT, a post-hoc framework that integrates conformal prediction with dynamic repairs to enforce fairness in computer vision systems, without retraining or modifying model parameters. Our method defines a fairness-aware non-conformity score incorporating both predictive error and demographic disparities and then uses a conformal threshold to ensure that only a controlled fraction of outputs violate fairness criteria. When a sample score exceeds this threshold, FAIR-SIGHT applies targeted corrections. Extensive experiments showed that FAIR-SIGHT significantly reduces unfairness while preserving accuracy across various tasks and backbones. These findings highlight its potential as a scalable, black-box solution for bias mitigation in real-world vision deployments. Future work will explore extending this approach to additional modalities, multi-attribute fairness, and more complex output structures.

{
    \small
    \bibliographystyle{ieeenat_fullname}
    \bibliography{main}
}

\end{document}